\documentclass[sigconf]{acmart}
\AtBeginDocument{%
  }

\usepackage{booktabs}
\usepackage{amsmath}
\usepackage{algorithm}
\usepackage{algorithmic}
\usepackage{multirow}
\usepackage{xcolor}
\usepackage{graphicx}
\usepackage{subcaption}
\usepackage{enumitem}
\usepackage{tikz}
\usepackage{makecell}
\usepackage{balance}
\usetikzlibrary{arrows.meta,positioning,calc,backgrounds,fit,shapes.geometric}

\newcommand{\ours}{\textrm{EVOL}}
\newcommand{\ep}{\mathrm{EP}}

\newcommand{\bc}{\mathrm{BC}}
\newcommand{\ppo}{\mathrm{PPO}}

\newcommand{\E}{\mathbb{E}}

\setcopyright{acmlicensed}
\copyrightyear{2026}
\acmYear{2026}
\setcopyright{cc}
\setcctype{by}
\acmConference[CIKM '26]{Proceedings of the 35th ACM International Conference on Information and Knowledge Management}{November 07--11, 2026}{Rome, Italy}
\acmBooktitle{Proceedings of the 35th ACM International Conference on Information and Knowledge Management (CIKM '26), November 07--11, 2026, Rome, Italy}
\acmDOI{10.1145/3799682.3841077}
\acmISBN{979-8-4007-2539-5/2026/11}

\begin{document}
\setlist[itemize]{leftmargin=10pt}
\setlist[enumerate]{leftmargin=15pt}

\title{EVOL: Simulator-Guided Evolutionary Expert Synthesis for Deployment-Free Learning Path Recommendation}

\author{Geonwoo Bang}
\orcid{0009-0005-4953-7321}
\affiliation{
\institution{Department of Computer Education, \\Sungkyunkwan University}
  \city{Seoul}
  \country{Republic of Korea}}
\email{g7199@g.skku.edu}

\author{Dongho Kim}
\orcid{0009-0004-3745-2438}
\affiliation{\institution{Department of Immersive Media Engineering/Social Innovation Convergence Program, \\Sungkyunkwan University} 
  \city{Seoul}
  \country{Republic of Korea}}
\email{dpl9753@g.skku.edu}

\author{Moohong Min}
\orcid{0000-0001-8979-1344}
\authornote{Corresponding author}
\affiliation{\institution{Department of Computer Education/Social Innovation Convergence Program, \\Sungkyunkwan University}   
\city{Seoul}
  \country{Republic of Korea}}
\email{iceo@skku.edu}

\renewcommand{\shortauthors}{Geonwoo Bang, Dongho Kim, and Moohong Min}

\begin{abstract}
Reinforcement learning (RL) for learning path recommendation (LPR) faces two coupled obstacles. First, the policy must commit to a sequence of $L$ concepts without intermediate feedback, producing a combinatorial search space that grows super-exponentially with $L$ and provides reward only at the final step. Second, expert learning paths would be the natural cure for sparse-reward RL, but they do not exist in educational data, because student logs record what learners did, not what they should have done. We address both obstacles by importing a recipe from simulator-based demonstration learning in robotics: the knowledge tracing simulator is used both to synthesize per-learner expert demonstrations through evolutionary search and to train a deployment-free policy that distills these demonstrations into a feed-forward learner. Our framework \ours{} instantiates this pipeline with an asymmetric actor-critic where the actor commits to deployment-realistic blind planning while the critic exploits the privileged simulator state during training. Across three datasets (ASSIST15, Junyi, EdNet; 39--189 concepts) and path lengths $L\in\{5,10,20\}$, \ours{} surpasses 8 baselines spanning heuristic, sequential, RL, graph-enhanced RL, and LLM-enhanced methods. We further compare three imitation strategies (BC, AWR, DAPG) and show that final performance is governed by the quality of evolutionary experts rather than by the particular imitation objective.
\end{abstract}

\begin{CCSXML}
<ccs2012>
<concept><concept_id>10002951.10003317.10003347.10003350</concept_id><concept_desc>Information systems~Recommender systems</concept_desc><concept_significance>500</concept_significance></concept>
<concept><concept_id>10010147.10010178.10010179</concept_id><concept_desc>Computing methodologies~Reinforcement learning</concept_desc><concept_significance>500</concept_significance></concept>
</ccs2012>
\end{CCSXML}
\ccsdesc[500]{Information systems~Recommender systems}
\ccsdesc[500]{Computing methodologies~Reinforcement learning}

\keywords{Sequential Recommendation, Learning Path Recommendation, Reinforcement Learning, Evolutionary Algorithms, Behavioral Cloning, Simulator-based Recommendation}


\maketitle

%% ============================================================
\section{Introduction}
%% ============================================================

Learning path recommendation (LPR) aims to recommend a personalized sequence of knowledge concepts that maximizes a learner's mastery of target concepts~\cite{liu2019cseal,li2023gehrl,cheng2025knowlp}. 
Given a curriculum of $M$ knowledge concepts, the goal is to select and order $L$ of them for an individual learner. 
Unlike next-item recommendation where each suggestion has independent utility, LPR requires joint optimization of the $L$-step sequence, since prerequisite dependencies, knowledge transfer, and cognitive load make the \emph{order} of practice 
critical~\cite{chen2023src,sweller1988cognitive,sweller1998cognitive}. Existing approaches frame LPR as a reinforcement learning (RL) problem inside a Knowledge Evolution Simulator (KES), a deep knowledge 
tracing (DKT)~\cite{piech2015dkt} model that predicts how mastery evolves with practice~\cite{liu2019cseal,chen2023src,li2023gehrl,cheng2025knowlp}.

However, this setup suffers from two coupled difficulties that prior work has only partially addressed.
First, the policy must perform blind sequential planning: the deployed system cannot ask a student to practice an intermediate concept just to peek at the resulting mastery, so the policy must commit to an entire $L$-step path from the initial mastery $\mathbf{h}_0$ alone. The combinatorial search space ($\binom{M}{L}\cdot L!$, reaching $1.2\!\times\!10^{45}$ for $M=189$ and $L=20$) combined with reward sparsity (the simulator returns mastery improvement only at step $L$) makes randomly initialized RL extremely sample-inefficient. Methods that bypass this by querying the simulator at every inference step~\cite{liu2019cseal,li2023gehrl,cheng2025knowlp} change the problem from deployment-time path recommendation to simulator-mediated online planning. In a real deployment, the learner history contains only interactions observed before the recommendation begins; however, step-wise KES planning repeatedly appends simulated responses to concepts the learner has not yet practiced, updates the mastery state from this simulator-augmented history, and then selects the next concept from that updated state. The resulting policy is therefore evaluated with access to counterfactual intermediate feedback that a deployable LPR system would not possess, making the setting mismatched with one-shot path recommendation~\cite{li2024privileged}.
Second, the natural cure for sparse-reward RL learning from expert demonstrations~\cite{rajeswaran2018dexterous,nair2018demonstrations,peng2019awr} is not available in education: student logs record what learners did, not what they should have done, and collecting optimal trajectories from human tutors at per-learner granularity is infeasible.

We resolve both difficulties by adapting the simulator-based demonstration learning recipe from robotics~\cite{rajeswaran2018dexterous,nair2018demonstrations} to educational recommendation. In robotic sim-to-real, policies are trained inside physics simulators such as MuJoCo~\cite{todorov2012mujoco} and Isaac Gym~\cite{makoviychuk2021isaac}, where expert trajectories can be generated, and then distilled into real-robot policies \cite{peng2018sim,hwangbo2019learning,yamada2024twist}. We observe a direct structural analogy: the KES plays the role of the physics engine, the learner is the ``robot,'' and a learning path is the ``trajectory.'' This analogy supplies the missing piece. Evolutionary search inside the KES synthesizes per-learner expert trajectories that human tutors cannot provide. Because the KES is derivative-free, cheap to query, and embarrassingly parallelizable~\cite{lange2023evosax,salimans2017es}, an evolutionary algorithm finds high-quality paths for thousands of learners in minutes per dataset at a scale impossible with online RL alone. These synthesized experts then bootstrap a simulator-free policy through imitation followed by RL fine-tuning, mirroring the BC to RL pipeline used in robotic manipulation~\cite{rajeswaran2018dexterous}.

Our framework \ours{} instantiates this pipeline with an explicit separation between simulator use during training and deployment-realistic execution. An asymmetric actor-critic architecture has the actor commit to blind planning from $\mathbf{h}_0$ in keeping the inference constraint while the critic exploits the privileged evolving mastery $\mathbf{h}_t$ for accurate value targets during training. After deployment, the actor alone produces complete paths in a single forward pass, without simulator dependency. We show that the contribution lies in the evolutionary expert synthesis rather than in any specific downstream learner: three imitation strategies (BC, AWR~\cite{peng2019awr}, DAPG~\cite{rajeswaran2018dexterous}) yield essentially the same final EP, while replacing evolutionary experts with random-path experts collapses performance.

Our contributions are:
\begin{itemize}
    \item \textbf{A deployment-free formulation of simulator-based LPR.}
    We frame LPR as demonstration learning from a knowledge evolution simulator, while requiring the deployed policy to recommend complete paths without simulator feedback.

    \item \textbf{EVOL: per-learner evolutionary expert synthesis.}
    We synthesize a distinct expert path for each learner inside the KES and distill these demonstrations into a simulator-free actor with asymmetric actor-critic refinement, turning sparse-reward path optimization into imitation-guided policy learning.

    \item \textbf{Mechanism-level empirical validation.}
    Experiments across three datasets show consistent gains over diverse baselines, and ablations isolate evolutionary expert quality and order-sensitive sequencing as the source of improvement.
\end{itemize}

%% ============================================================
\section{Related Work}
%% ============================================================

\subsection{Knowledge Tracing}

Knowledge tracing (KT) models estimate a learner's evolving mastery over time from interaction sequences \cite{corbett1994knowledge,shen2024survey}. Deep Knowledge Tracing (DKT)~\cite{piech2015dkt} pioneered the use of recurrent neural networks for this task. Subsequent models include DKVMN~\cite{zhang2017dkvmn}, SAKT~\cite{pandey2019sakt}, AKT~\cite{ghosh2020akt}, and SAINT~\cite{choi2020saint}, which leverage attention mechanisms and memory networks for improved accuracy. In LPR, KT models serve dual roles: as the knowledge evolution simulator (KES) for RL training and as the mastery estimator at inference time \cite{liu2019cseal}.

\subsection{Learning Path Recommendation}

Early LPR methods relied on heuristic rules such as prerequisite ordering~\cite{chen2018prerequisite} and difficulty-aware recommendation~\cite{liu2019cseal}. 
CSEAL~\cite{liu2019cseal} introduced the KES-based RL paradigm, training an actor-critic recommender within a DKT simulator and using cognitive navigation to guide concept selection. 
Subsequent KES-based methods include GEHRL~\cite{li2023gehrl}, which incorporates hierarchical goal planning and graph-based candidate filtering, and DLELP/KnowLP~\cite{cheng2025knowlp}, which use prerequisite and similarity structures for reinforcement-learning-based path generation. 
In parallel, sequential recommendation architectures such as GRU4Rec~\cite{hidasi2016gru4rec} and SASRec~\cite{kang2018sasrec} provide natural baselines for generating ordered learning paths.

\subsection{Sim-to-Real and Demonstration-Guided RL}

In robotic manipulation, the dominant strategy for sparse-reward, long-horizon tasks is to (i) generate expert trajectories inside a physics simulator, (ii) pre-train a policy with behavioral cloning (BC), and (iii) refine it with RL while anchoring on the demonstrations~\cite{rajeswaran2018dexterous,nair2018demonstrations}. DAPG~\cite{rajeswaran2018dexterous} augments the natural policy gradient with a decaying BC term so the policy stays close to demonstrations early in training; AWR~\cite{peng2019awr} weights expert state--action pairs by their advantage, allowing off-policy demonstrations of varying quality to contribute proportionally to their estimated value. Both methods assume access to demonstrations collected by scripted controllers, motion capture, or teleoperation. The same assumption underlies imitation learning surveys and benchmarks~\cite{ross2011dagger,torabi2018bc}.

This assumption breaks in educational recommendation: there is no oracle tutor who has solved each learner's path. While prior LPR work uses the KES only as a training environment~\cite{liu2019cseal}, our contribution is to identify that it can also serve as the source of demonstrations through evolutionary search. The downstream pipeline (BC, AWR, or DAPG fine-tuning) is then standard sim-to-real machinery transplanted to a non-robotic domain.

\subsection{Evolutionary Search for RL Initialization}

Evolutionary algorithms (EAs) are well suited to sparse-reward combinatorial problems: they evaluate complete solutions end-to-end and select on fitness alone, so they do not require informative gradient signals~\cite{salimans2017es}. Fitness evaluations are also independent, which makes them embarrassingly parallel on GPUs~\cite{lange2023evosax}. In our setting, a population of 60 candidate paths can be batch-evaluated in a single DKT forward pass. Hybrid evolutionary RL methods~\cite{li2024evorl,khadka2018erl,jaderberg2017pbt} typically evolve policy parameters alongside gradient-trained actors, sharing replay buffers between the two populations.

\ours{} differs in granularity: we evolve individual solutions (per-learner paths) rather than policy parameters, and we use the evolved solutions only as demonstrations for a separately trained neural policy. Evolution thus operates at the instance level (one search per learner) while gradient-based learning operates at the population level (one policy generalizing across learners). This separation is essential because at deployment we cannot run an evolutionary search per new learner (the simulator is unavailable), so we must distill the evolved solutions into a feed-forward policy.

%% ============================================================
\section{Preliminaries}
%% ============================================================

\subsection{Problem Formulation}

%Let $\mathcal{K} = \{k_1, k_2, \ldots, k_M\}$ denote a set of $M$ knowledge concepts in a curriculum. A learner is characterized by their interaction history $\mathcal{H} = \{(c_1, r_1), (c_2, r_2), \ldots\}$ where $c_t \in \mathcal{K}$ is the practiced concept and $r_t \in \{0, 1\}$ is the binary correctness outcome at step $t$.

%Given a learner's history $\mathcal{H}$, a set of target concepts $\mathcal{G} \subseteq \mathcal{K}$, and a path length $L$, the goal is to find an ordered sequence $\mathbf{p} = (p_1, p_2, \ldots, p_L) \in \mathcal{K}^L$ that maximizes the learner's mastery improvement on the targets $\mathcal{G}$.

Let $\mathcal{K} = \{k_1, k_2, \ldots, k_M\}$ denote a set of $M$ knowledge concepts in a curriculum. 
A learner is characterized by an interaction history $\mathcal{H} = \{(c_1, r_1), (c_2, r_2), \ldots\}$, where $c_t \in \mathcal{K}$ is the practiced concept and $r_t \in \{0,1\}$ is the binary correctness outcome at step $t$. 
Given $\mathcal{H}$, a set of target concepts $\mathcal{G} \subseteq \mathcal{K}$, and a path length $L$, learning path recommendation seeks an ordered path $\mathbf{p}=(p_1,\ldots,p_L)$ that maximizes the learner's mastery improvement on $\mathcal{G}$ after the path is executed. 
Throughout this paper, we consider non-repeating paths, i.e., $p_t \in \mathcal{K}$ and $p_t \neq p_{t'}$ for $t \neq t'$, which matches the action masking and search-space analysis used in our method. 
Under the deployment-free setting studied here, the policy must generate the complete path from the learner's initial mastery estimate and target concepts, without using the KES to simulate intermediate responses or update mastery during path generation.

\subsection{Knowledge Evolution Simulator (KES)}

Following the CSEAL framework~\cite{liu2019cseal}, we employ a trained DKT model~\cite{piech2015dkt} as the KES. Given a history $\mathcal{H}$, the DKT produces a mastery vector $\mathbf{h} \in [0,1]^M$ where $h_i$ represents the estimated probability of correctly answering concept $k_i$.

The KES operates with a deterministic response model, following the EduSim/CSEAL evaluation protocol~\cite{liu2019cseal}: when a learner practices concept $c$, the outcome is determined by
\begin{equation}
    r = \mathbf{1}[h_c > 0.5],
\end{equation}
where $h_c$ is the learner's current mastery of concept $c$.\footnote{This deterministic thresholding is the default of the official EduSim simulator (\texttt{KSSScorer} with \texttt{binary\_scorer=True}) and is adopted by subsequent KES-based LPR work~\cite{li2023gehrl,cheng2025knowlp}. A stochastic alternative ($r\!\sim\!\mathrm{Bernoulli}(h_c)$) exists in adjacent exercise-recommendation literature; we adopt the deterministic protocol for direct comparability with our baselines, and apply the same rule uniformly to all methods during both training and evaluation.} The interaction $(c, r)$ is appended to $\mathcal{H}$, and the DKT produces an updated mastery vector.

\subsection{Effectiveness Percentage (EP)}

The standard evaluation metric for LPR is the Effectiveness Percentage~\cite{liu2019cseal,cheng2025knowlp}, which measures the normalized mastery improvement on target concepts after executing a learning path:
\begin{equation}
    \ep = \frac{\sum_{g \in \mathcal{G}} \bigl(E_g^{\text{end}} - E_g^{\text{start}}\bigr)}{\sum_{g \in \mathcal{G}} \bigl(1 - E_g^{\text{start}}\bigr)},
    \label{eq:ep}
\end{equation}
where $E_g^{\text{start}}$ and $E_g^{\text{end}}$ are the learner's mastery of target $g$ before and after the path. This aggregate form~\cite{cheng2025knowlp} naturally down-weights near-ceiling targets by weighting each target proportionally to its room for improvement $(1 - E_g^{\text{start}})$.

\subsection{Concept Prerequisite Graph}

We construct a concept graph $\mathcal{G}_c = (\mathcal{K}, \mathcal{E}_p, \mathcal{E}_s)$ where $\mathcal{E}_p$ denotes prerequisite edges and $\mathcal{E}_s$ denotes similarity edges. Following prior work~\cite{liu2019cseal,cheng2025knowlp}, we derive these from the DKT's cross-concept influence: for each concept $k_i$, we measure the change in mastery of all other concepts when $k_i$ is practiced correctly. An edge $(k_i, k_j) \in \mathcal{E}_p$ if the influence exceeds a threshold $\tau$.

%% ============================================================
\section{Method}

\begin{figure*}[t]
 \centering
 \includegraphics[width=\textwidth]{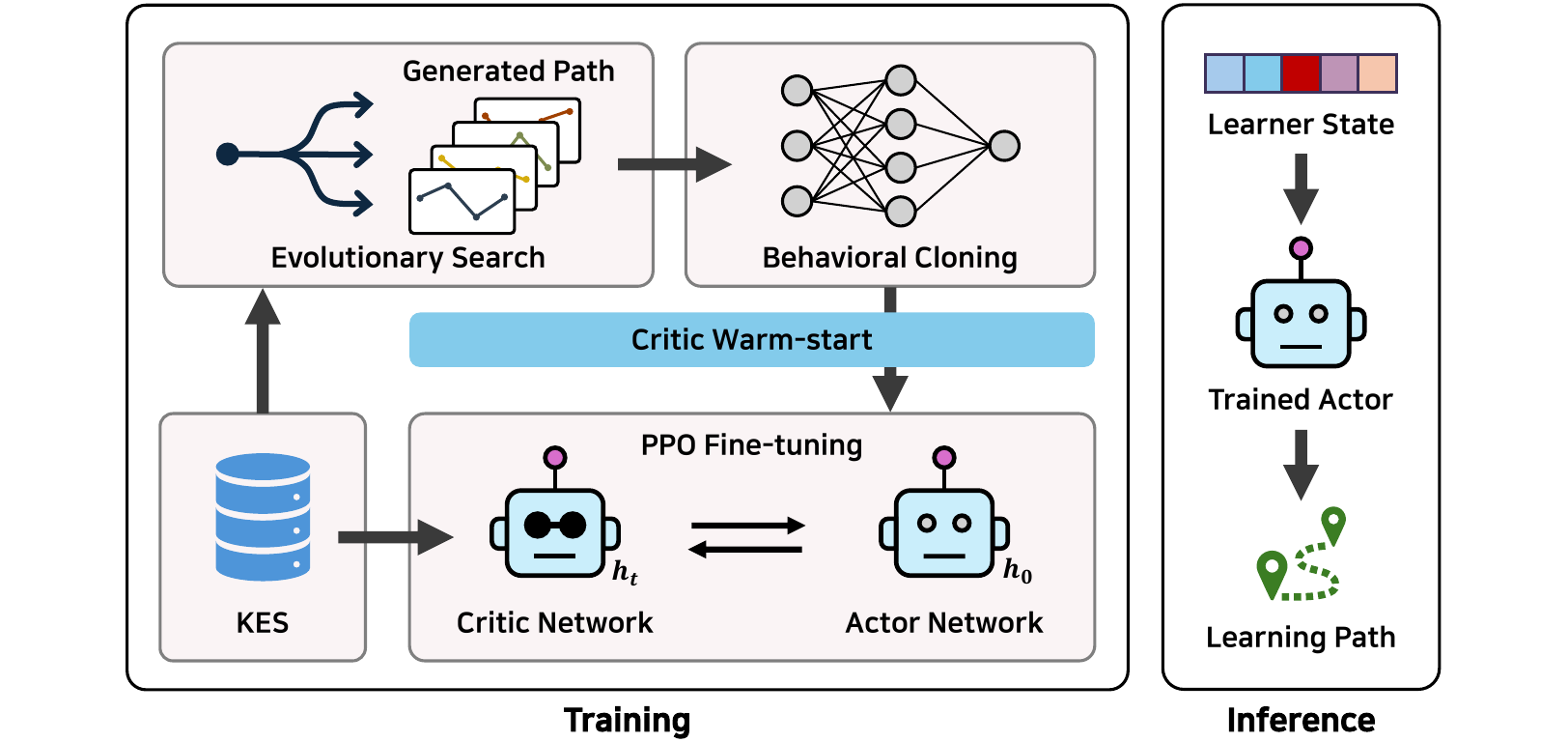} % 도식 파일 경로
 \caption{Overview of \ours{}. The KES is consulted only at train time.
\textbf{Stage~1} runs evolutionary search inside the KES to synthesize
per-learner expert paths; \textbf{Stage~2} distills them into a feed-forward
actor conditioned only on the initial mastery $\mathbf{h}_0$;
\textbf{Stage~2.5} warm-starts the critic on the privileged $\mathbf{h}_t$;
\textbf{Stage~3} refines the actor with asymmetric PPO. At deployment the
actor alone emits the full $L$-step path with no KES query.}
 \label{fig:framework_overview}
\end{figure*}

\ours{} has three training stages (Figure~\ref{fig:framework_overview}). All three rely on the KES for evaluation, but the deployed policy is simulator-free: only the actor network is shipped, and at inference it repeatedly consumes the initial mastery $\mathbf{h}_0$, target mask $\mathbf{g}$, step index, and used-concept mask to emit an $L$-step path, without querying the KES.

\subsection{Stage 1: Evolutionary Expert Generation}
\label{sec:method_stage1}

For each training learner with history $\mathcal{H}$, targets $\mathcal{G}$, and initial mastery $\mathbf{h}_0$, we search for high-quality learning paths using an evolutionary algorithm.

\textbf{Candidate Space.} We construct a candidate set $\mathcal{C}$ of size at most $C$ by combining the target concepts, their prerequisites and similar concepts from the concept graph, and zone-of-proximal-development (ZPD) concepts whose mastery is closest to 0.5:
\begin{equation}
    \mathcal{C} = \mathcal{G} \cup \bigcup_{g \in \mathcal{G}} \left(\text{prereq}(g) \cup \text{sim}_K(g)\right) \cup \text{ZPD}.
\end{equation}
Here, $\text{prereq}(g)$ and $\text{sim}_K(g)$ are defined on the concept graph $\mathcal{G}_c$: $\text{prereq}(g)$ denotes the prerequisite neighbors of target $g$, and $\text{sim}_K(g)$ denotes its top-$K$ similarity neighbors. 
$\text{ZPD}$ denotes the set of concepts whose current mastery under $\mathbf{h}_0$ is closest to $0.5$, approximating the learner's zone of proximal development.

\textbf{Evolutionary Search.} We maintain a population of $N_{\text{pop}}$ candidate paths, each of length $L$ drawn from $\mathcal{C}$. Over $N_{\text{gen}}$ generations, the population evolves through tournament selection, ordered crossover, and point mutation. The fitness of each path $\mathbf{p}$ is evaluated in the KES:
\begin{equation}
    f(\mathbf{p}) = \ep(\mathcal{H}, \mathbf{p}, \mathcal{G}).
\end{equation}

\textbf{Expert Selection.} From the final population, we select the single best path per learner ($K=1$), yielding one expert trajectory per training learner. We find that selecting the highest-fitness path produces cleaner training signal for BC compared to diverse multi-path selection: with $K>1$ paths per learner, the same state maps to multiple conflicting actions, creating label noise that caps BC accuracy at $\sim$70\%. With $K=1$, each state has a unique target action, enabling sharper policy initialization.

\subsection{Stage 2: Behavioral Cloning}

The evolutionary expert trajectories provide state--action pairs for supervised
learning. Consistent with the deployment-free constraint, BC conditions only on
$\mathbf{h}_0$, the target mask, and the step index, never on KES-updated
intermediate mastery. For each expert $(\mathbf{h}_0, \mathcal{G}, \mathbf{p}, \text{EP})$,
we construct $L$ training examples:
\begin{equation}
    \{(\mathbf{s}_t, a_t)\}_{t=0}^{L-1}, \quad \mathbf{s}_t = [\mathbf{h}_0 \| \mathbf{g} \| t/L],
\end{equation}
where $\mathbf{g} \in \{0,1\}^M$ is the target mask, $t/L$ provides step position, and $a_t = p_t$ is the expert's action at step $t$. The step index and the used-concept exclusion mask (applied during action selection) are the only signals that differentiate steps, so the actor learns to plan an entire path from the initial knowledge state alone.

The policy $\pi_\theta$ is trained to minimize the cross-entropy loss:
\begin{equation}
    \mathcal{L}_{\bc} = -\frac{1}{|\mathcal{D}|} \sum_{(\mathbf{s}, a) \in \mathcal{D}} \log \pi_\theta(a \mid \mathbf{s}).
\end{equation}
The BC optimizer updates only the actor parameters; the critic head is trained separately in Stage~2.5 because its target distribution (privileged $\mathbf{h}_t$) differs from the actor's deployment distribution (blind $\mathbf{h}_0$).

\subsection{Stage 2.5: Critic Warm-Start}
\label{sec:critic_ws}

The PPO critic estimates $V(\mathbf{s}_t^{\text{critic}})$ where $\mathbf{s}_t^{\text{critic}}$ contains the evolving mastery $\mathbf{h}_t$. Initializing this head from scratch wastes early PPO updates on critic learning before useful advantage estimates appear. We therefore warm-start the critic on the expert trajectories: for each expert path we roll out the KES, compute the discounted return $G_t = \gamma^{L-1-t}\,\text{EP}$ at every step, and fit $V$ by mean-squared regression to $G_t$. The actor parameters are frozen during this stage.

\subsection{Stage 3: PPO Fine-Tuning}

Starting from the BC-initialized policy, we apply PPO~\cite{schulman2017ppo} for online fine-tuning. At each episode, we sample a batch of learners and collect trajectories by interacting with the KES.

\textbf{Simulator-Free Inference via Asymmetric Actor-Critic.}

A critical design consideration in KES-based LPR is whether the policy may access the simulator at inference time. As noted by Li et al.~\cite{li2024privileged}, the simulator's internal knowledge state is unavailable in real deployment---a tutoring system cannot observe how a student's mastery evolves before they actually practice. Relying on per-step simulator feedback during inference is akin to planning a study schedule while already knowing the exam answers.

We address this through an asymmetric actor-critic architecture. The \textbf{actor} (policy) receives only the initial mastery $\mathbf{h}_0$ at every step, ensuring KES-free inference:
\begin{equation}
    \mathbf{s}_t^{\text{actor}} = [\mathbf{h}_0 \| \mathbf{g} \| t/L], \quad a_t \in \mathcal{K} \setminus \{a_0, \ldots, a_{t-1}\},
\end{equation}
where $\mathbf{g} \in \{0,1\}^M$ is the target mask and $t/L$ provides step position. The actor distinguishes steps through the step index and the used concept mask that excludes previously selected concepts from the action space. At deployment, the complete path is generated by $L$ forward passes through the actor alone, without simulator access.

The \textbf{critic} (value function), used only during training, receives the privileged evolving mastery $\mathbf{h}_t$ from the KES:
\begin{equation}
    \mathbf{s}_t^{\text{critic}} = [\mathbf{h}_t \| \mathbf{g} \| t/L].
\end{equation}
This asymmetry allows the critic to provide more accurate advantage estimates by observing how each recommendation affects mastery, while the actor learns to plan without this privileged information. The actor-critic separation ensures the deployed policy is fully simulator-free: it takes a learner's initial knowledge state and target concepts as input, and outputs a complete learning path through $L$ actor calls, without any KES query or simulated intermediate feedback.

\textbf{Reward.} We use a sparse terminal reward computed via KES simulation:
\begin{equation}
    R_t = \begin{cases} \ep(\mathcal{H}, \mathbf{p}_{0:L}, \mathcal{G}) & \text{if } t = L-1 \\ 0 & \text{otherwise} \end{cases}
\end{equation}

\textbf{Policy Optimization.} The PPO clipped surrogate objective is:
\begin{equation}
    \mathcal{L}_{\ppo} = -\E_t\left[\min\left(\rho_t \hat{A}_t,\ \text{clip}(\rho_t, 1\pm\epsilon) \hat{A}_t\right)\right],
\end{equation}
where $\rho_t = \pi_\theta(a_t|\mathbf{s}_t^{\text{actor}}) / \pi_{\theta_{\text{old}}}(a_t|\mathbf{s}_t^{\text{actor}})$ is the importance ratio and $\hat{A}_t = R_t + \gamma V(\mathbf{s}_{t+1}^{\text{critic}}) - V(\mathbf{s}_t^{\text{critic}})$ uses the privileged critic. The full loss includes a value function term and an entropy bonus:
\begin{equation}
    \mathcal{L} = \mathcal{L}_{\ppo} + c_v \cdot \mathcal{L}_{\text{value}} - c_e \cdot H[\pi_\theta].
\end{equation}

\subsection{Expert-Utilization Variants}
\label{sec:variants}

Stages~2 and~3 above describe \ours{}-BC, the canonical instantiation. The downstream learner that consumes the evolutionary experts is, however, a design choice. We therefore consider two additional variants drawn from the sim-to-real literature:

\textbf{\ours{}-AWR.} Following Peng et al.~\cite{peng2019awr}, we replace Stage~2 with advantage-weighted regression. Each expert sample $(\mathbf{s},a)$ is weighted by $\exp(\hat{A}(\mathbf{s},a)/\beta)$, where the advantage is computed against a jointly updated value baseline. AWR up-weights expert actions whose realized return exceeds the baseline, making it a natural choice when experts have heterogeneous quality.

\textbf{\ours{}-DAPG.} Following Rajeswaran et al.~\cite{rajeswaran2018dexterous}, we keep Stage~2 as BC but replace Stage~3 with PPO augmented by a decaying BC term: $\mathcal{L} = \mathcal{L}_{\ppo} + \lambda_k\,\mathcal{L}_{\bc}^{\text{demo}}$ with $\lambda_k = \lambda_0\, \lambda_1^k$. This keeps the policy tethered to demonstrations early in PPO and gradually releases it.

All three variants share Stage~1 and the critic warm-start, and differ only in how they use the synthesized experts during imitation or PPO refinement. 
This design lets us test whether performance depends on the downstream imitation objective or on the quality of the evolutionary experts themselves.

%% ============================================================
\section{Experiments}
%% ============================================================

\subsection{Datasets}

We evaluate on three educational datasets that span the typical concept-count range of LPR studies (39--189 KCs):

\begin{itemize}
    \item \textbf{ASSIST15}~\cite{feng2009assistments}: An online mathematics tutoring platform. After filtering learners with $\geq$10 interactions, it contains 14,567 learners across 100 skill-builder concepts and 682K interactions.
    \item \textbf{Junyi Academy}~\cite{chang2015junyi}: an online mathematics learning platform. We sample 20,000 learners (seed-controlled) over 39 knowledge concepts and 4.9M interactions.
    \item \textbf{EdNet-KT1}~\cite{choi2020ednet}: A large-scale English-learning dataset from Santa. We sample 50,000 learners over 189 knowledge tags (from the official EdNet-Contents metadata) and 7.5M interactions.
\end{itemize}

Following the CSEAL protocol~\cite{liu2019cseal}, each dataset is split 50/50 into $\mathcal{D}_{\text{Sim}}$ (DKT training) and $\mathcal{D}_{\text{Off}}$ (LPR training and evaluation); $\mathcal{D}_{\text{Off}}$ is further split 80/10/10 into training, validation, and test partitions. Splits are seed-controlled so the three random seeds we report induce different student populations.

\subsection{Baselines}

We compare against 8 baselines spanning four categories. 
For the heuristic category, Table~\ref{tab:main} reports \textit{Target-priority}, which first places the target concepts and then fills the remaining positions with graph-neighbor or ZPD concepts when $L>|\mathcal{G}|$. 
The remaining baselines cover sequential recommenders, RL from scratch, graph-enhanced RL, and LLM-enhanced LPR.

\begin{itemize}
    \item \textbf{Sequential models with the same supervision:} \textit{GRU4Rec}~\cite{hidasi2016gru4rec} and \textit{SASRec}~\cite{kang2018sasrec} are standard recurrent and transformer-based sequence models. We train them on the same evolutionary trajectories as \ours{}-BC so that they share the supervisory signal and differ only in the downstream learner; the comparison therefore isolates the contribution of RL fine-tuning beyond pure imitation.
    \item \textbf{RL from scratch:} \textit{PPO-vanilla}~\cite{schulman2017ppo} uses the same asymmetric actor-critic architecture as \ours{} but is initialized from random weights and receives no expert demonstrations. This isolates the contribution of evolutionary pre-exploration.
    \item \textbf{Graph-enhanced RL:} \textit{CSEAL}~\cite{liu2019cseal} (A2C with cognitive navigation), \textit{GEHRL}~\cite{li2023gehrl} (hierarchical RL with goal planning), and \textit{DLELP}~\cite{cheng2025knowlp} (PPO with a plateau-detection similarity agent and a DKT-derived concept graph).
    \item \textbf{LLM-enhanced:} \textit{KnowLP}~\cite{cheng2025knowlp} shares the DLELP architecture but replaces the DKT-derived concept graph with an LLM-constructed EDU-GraphRAG graph. The same P-Agent and graph-aware candidate mask are used at training and inference, so the only difference between DLELP and KnowLP is the source of the graph.
\end{itemize}

We also report three expert-utilization variants of our method: \textit{\ours{}-BC}, \textit{\ours{}-AWR}, and \textit{\ours{}-DAPG} (Section~\ref{sec:variants}). 
All three share Stage~1 and differ only in how they exploit the synthesized experts downstream. All baselines and variants are re-implemented in a unified codebase. Table~\ref{tab:baseline_fidelity} summarizes which aspects of each baseline are preserved verbatim from the original publication and which are modified for fair comparison. Specifically, we replace inference-time KES queries with $\mathbf{h}_0$-only planning so that all methods produce deployment-realistic recommendations, and we equalize the policy network capacity across PPO-based methods (asymmetric actor-critic with hidden 256/512 for Junyi/ASSIST15--EdNet). We emphasize the scope of this comparison: we do not claim superiority over the original, inference-time simulator-querying versions of CSEAL/GEHRL/DLELP/KnowLP under their own assumptions. Rather, we compare all methods under the deployment-free constraint targeted by this paper---namely, that no KT query is available at inference time on the live student. The performance gap observed under this constraint is precisely the quantity practitioners face when deploying any of these methods without a runtime KES. $\mathbf{h}_0$-only inference is the deployment definition: a learner's future mastery is unknowable before practice, so any method reported under inference-time KES access is measuring a counterfactual oracle rather than a servable system.

\begin{table*}[t]
\setlength{\tabcolsep}{2pt}
\centering
\caption{Baseline fidelity. ``Original'' columns preserve the published method; ``Adapted'' columns describe our changes for deployment-realistic, capacity-matched comparison.}
\label{tab:baseline_fidelity}
\small
\begin{tabular}{@{}l|cccc@{}}
\toprule
Method & Algorithm preserved & Reward preserved & \makecell{Inference uses KES\\(original)} & \makecell{Inference uses KES\\(ours)} \\
\midrule
GRU4Rec~\cite{hidasi2016gru4rec}    & yes (GRU)                 & N/A (BC on experts)       & no  & no \\
SASRec~\cite{kang2018sasrec}        & yes (transformer)         & N/A (BC on experts)       & no  & no \\
PPO-vanilla~\cite{schulman2017ppo}  & yes (PPO + clip)          & terminal EP               & N/A & no (asymmetric AC) \\
CSEAL~\cite{liu2019cseal}           & yes (A2C + cognitive nav) & EP                        & yes & no (h$_0$-only actor) \\
GEHRL~\cite{li2023gehrl}            & yes (hierarchical PPO)    & EP                        & yes & no (h$_0$-only actor) \\
DLELP~\cite{cheng2025knowlp}        & yes (PPO + plateau S-agent)& EP                        & yes & no (S-agent off in h$_0$ mode; graph-aware candidate mask retained) \\
KnowLP~\cite{cheng2025knowlp}       & yes (DLELP + LLM graph)   & EP                        & yes & no (graph is EDU-GraphRAG instead of DKT-derived) \\
\bottomrule
\end{tabular}
\end{table*}

\subsection{Implementation Details}
All methods are implemented in a unified experimental pipeline, with shared simulator construction, evaluation protocol, and training budget whenever applicable.\footnote{All scripts are available at \url{https://github.com/g7199/EvoLearning}.}

\textbf{KES.} 
We train the KES as a DKT model~\cite{piech2015dkt} with an LSTM backbone on $\mathcal{D}_{\text{Sim}}$, using the pyKT framework~\cite{liu2022pykt} and its standard configuration. 
The DKT uses embedding size 200, dropout 0.1, learning rate $10^{-3}$, and batch size 256. 
The resulting validation AUCs are 0.70 on ASSIST15, 0.74 on Junyi, and 0.76 on EdNet.

\textbf{Concept Graph.} 
We derive prerequisite and similarity graphs from the DKT's cross-concept influence matrix. 
The influence threshold $\tau$ is set to 0.25 for ASSIST15, 0.05 for Junyi, and 0.15 for EdNet, with thresholds chosen so that the average degree is comparable across datasets.

\textbf{Evolutionary Search.} 
For evolutionary search, we use population size $N_{\text{pop}}=60$, number of generations $N_{\text{gen}}=50$, and candidate cap $C=30$. 
We retain the single best path per learner ($K=1$), and discuss the implications of using $K>1$ in Section~\ref{sec:method_stage1}. 
The wall-clock generation time on a single GPU is 8.4 minutes for Junyi (8,000 learners), 31 minutes for ASSIST15 (5,827 learners), and 41 minutes for EdNet (19,880 learners).

\textbf{Policy Network.} 
We use an asymmetric actor-critic architecture with two separate two-layer MLPs. 
The hidden width is 512 for ASSIST15 and EdNet, and 256 for Junyi. 
The actor consumes $[\mathbf{h}_0\,\|\,\mathbf{g}\,\|\,t/L]$, while the critic consumes $[\mathbf{h}_t\,\|\,\mathbf{g}\,\|\,t/L]$.

\textbf{Training.} 
We train the actor with behavioral cloning for 2,000 epochs using Adam with learning rate $10^{-3}$, weight decay $10^{-4}$, and batch size 4096. 
The critic warm-start stage uses 1,000 epochs of mean-squared error regression on expert returns. 
For PPO fine-tuning, we train for 30,000 episodes with batch size 128, learning rate $10^{-4}$, clip parameter $\epsilon=0.15$, entropy coefficient 0.03, a cosine learning-rate schedule, 4 inner PPO epochs per batch, and gradient clipping at 0.5. 
PPO-vanilla uses the same hyperparameters but no BC initialization. 
All experiments are repeated with three seeds ($\{42,123,7\}$), and we report mean$\pm$std throughout. 
Experiments are run on a workstation with an Intel Xeon E5-2620 v4 CPU and a single NVIDIA GeForce RTX 3090 GPU.

\subsection{Main Results}

Table~\ref{tab:main} presents the EP scores across all datasets and path lengths.

\textbf{Consistent gains under deployment-free inference.}
At least one \ours{} variant achieves the best result in every dataset--length setting, and the three \ours{} variants remain tightly clustered throughout. The strongest non-\ours{} baseline is usually PPO-vanilla, which shares the same asymmetric actor-critic architecture, training budget, and reward definition, but lacks evolutionary expert pretraining. The consistent advantage over this baseline indicates that the gain is not primarily due to architecture or optimizer choice, but to the expert synthesis stage. The advantage is especially clear in longer-path and larger-concept settings, where sparse terminal reward and combinatorial action spaces make exploration harder.

%The robustness analyses (Sections~\ref{sec:cross_sim} and~\ref{sec:stochastic}) further show that the within-simulator gain is amplified under DKT-instance shift ($+0.071\!\to\!+0.089$) and preserved under all three stochastic response models---indicating that the gain reflects a genuinely better policy rather than a deterministic-single-DKT artifact.

\begin{table*}[t]
\centering
\caption{Test EP (mean{\scriptsize$\pm$std} over 3 seeds) across datasets and path lengths. Best in \textbf{bold}, second best \underline{underlined}.}
\label{tab:main}
\resizebox{\textwidth}{!}{%
\begin{tabular}{@{}cl|c|cc|c|ccc|c|ccc@{}}
\toprule
& & Heuristic & \multicolumn{2}{c|}{Sequential} & RL & \multicolumn{3}{c|}{Graph-enhanced RL} & LLM & \multicolumn{3}{c}{\textbf{Ours}} \\
Dataset & $L$ & Target & GRU4Rec & SASRec & PPO & CSEAL & GEHRL & DLELP & KnowLP & \ours{}-BC & \ours{}-AWR & \ours{}-DAPG \\
\midrule
\multirow{3}{*}{\shortstack[c]{\textbf{Junyi}\\{\scriptsize 39 KCs}}}
& 5  & +0.203{\scriptsize$\pm$.003} & +0.237{\scriptsize$\pm$.008} & +0.256{\scriptsize$\pm$.007} & +0.247{\scriptsize$\pm$.010} & +0.054{\scriptsize$\pm$.049} & +0.106{\scriptsize$\pm$.003} & +0.110{\scriptsize$\pm$.013} & +0.089{\scriptsize$\pm$.019} & \underline{+0.286}{\scriptsize$\pm$.003} & \textbf{+0.286}{\scriptsize$\pm$.000} & +0.285{\scriptsize$\pm$.005} \\
& 10 & +0.244{\scriptsize$\pm$.002} & +0.266{\scriptsize$\pm$.009} & +0.320{\scriptsize$\pm$.006} & +0.364{\scriptsize$\pm$.003} & +0.178{\scriptsize$\pm$.008} & +0.138{\scriptsize$\pm$.008} & +0.247{\scriptsize$\pm$.003} & +0.230{\scriptsize$\pm$.010} & \textbf{+0.390}{\scriptsize$\pm$.001} & +0.388{\scriptsize$\pm$.001} & \underline{+0.390}{\scriptsize$\pm$.001} \\
& 20 & +0.286{\scriptsize$\pm$.001} & $-$0.005{\scriptsize$\pm$.017} & +0.299{\scriptsize$\pm$.006} & +0.380{\scriptsize$\pm$.005} & +0.207{\scriptsize$\pm$.006} & +0.045{\scriptsize$\pm$.029} & +0.337{\scriptsize$\pm$.010} & +0.337{\scriptsize$\pm$.010} & \underline{+0.426}{\scriptsize$\pm$.002} & +0.420{\scriptsize$\pm$.002} & \textbf{+0.428}{\scriptsize$\pm$.002} \\
\midrule
\multirow{3}{*}{\shortstack[c]{\textbf{ASSIST15}\\{\scriptsize 100 KCs}}}
& 5  & +0.296{\scriptsize$\pm$.011} & +0.329{\scriptsize$\pm$.007} & +0.355{\scriptsize$\pm$.008} & +0.420{\scriptsize$\pm$.011} & +0.062{\scriptsize$\pm$.022} & +0.211{\scriptsize$\pm$.007} & +0.251{\scriptsize$\pm$.015} & +0.189{\scriptsize$\pm$.034} & +0.470{\scriptsize$\pm$.003} & \underline{+0.471}{\scriptsize$\pm$.005} & \textbf{+0.472}{\scriptsize$\pm$.004} \\
& 10 & +0.321{\scriptsize$\pm$.009} & +0.325{\scriptsize$\pm$.010} & +0.466{\scriptsize$\pm$.011} & +0.543{\scriptsize$\pm$.009} & +0.215{\scriptsize$\pm$.013} & +0.228{\scriptsize$\pm$.003} & +0.454{\scriptsize$\pm$.002} & +0.388{\scriptsize$\pm$.011} & \textbf{+0.614}{\scriptsize$\pm$.002} & +0.604{\scriptsize$\pm$.001} & \underline{+0.613}{\scriptsize$\pm$.003} \\
& 20 & +0.331{\scriptsize$\pm$.007} & +0.332{\scriptsize$\pm$.007} & +0.473{\scriptsize$\pm$.005} & +0.511{\scriptsize$\pm$.009} & +0.292{\scriptsize$\pm$.005} & +0.142{\scriptsize$\pm$.018} & +0.230{\scriptsize$\pm$.015} & +0.293{\scriptsize$\pm$.062} & \underline{+0.621}{\scriptsize$\pm$.000} & +0.619{\scriptsize$\pm$.003} & \textbf{+0.624}{\scriptsize$\pm$.001} \\
\midrule
\multirow{3}{*}{\shortstack[c]{\textbf{EdNet}\\{\scriptsize 189 KCs}}}
& 5  & +0.104{\scriptsize$\pm$.001} & +0.188{\scriptsize$\pm$.001} & +0.200{\scriptsize$\pm$.000} & +0.263{\scriptsize$\pm$.003} & +0.136{\scriptsize$\pm$.006} & +0.052{\scriptsize$\pm$.002} & +0.197{\scriptsize$\pm$.006} & +0.165{\scriptsize$\pm$.003} & \underline{+0.277}{\scriptsize$\pm$.002} & +0.276{\scriptsize$\pm$.002} & \textbf{+0.278}{\scriptsize$\pm$.002} \\
& 10 & +0.138{\scriptsize$\pm$.001} & +0.240{\scriptsize$\pm$.005} & +0.260{\scriptsize$\pm$.003} & +0.367{\scriptsize$\pm$.003} & +0.187{\scriptsize$\pm$.006} & +0.045{\scriptsize$\pm$.009} & +0.264{\scriptsize$\pm$.009} & +0.227{\scriptsize$\pm$.002} & \underline{+0.383}{\scriptsize$\pm$.003} & +0.379{\scriptsize$\pm$.003} & \textbf{+0.384}{\scriptsize$\pm$.003} \\
& 20 & +0.172{\scriptsize$\pm$.002} & +0.266{\scriptsize$\pm$.006} & +0.275{\scriptsize$\pm$.008} & +0.457{\scriptsize$\pm$.004} & +0.229{\scriptsize$\pm$.007} & $-$0.089{\scriptsize$\pm$.006} & +0.352{\scriptsize$\pm$.010} & +0.321{\scriptsize$\pm$.004} & \textbf{+0.475}{\scriptsize$\pm$.003} & +0.466{\scriptsize$\pm$.005} & \underline{+0.474}{\scriptsize$\pm$.004} \\
\bottomrule
\end{tabular}%
}
\end{table*}

\begin{figure}[t]
    \centering
    \includegraphics[width=\columnwidth]{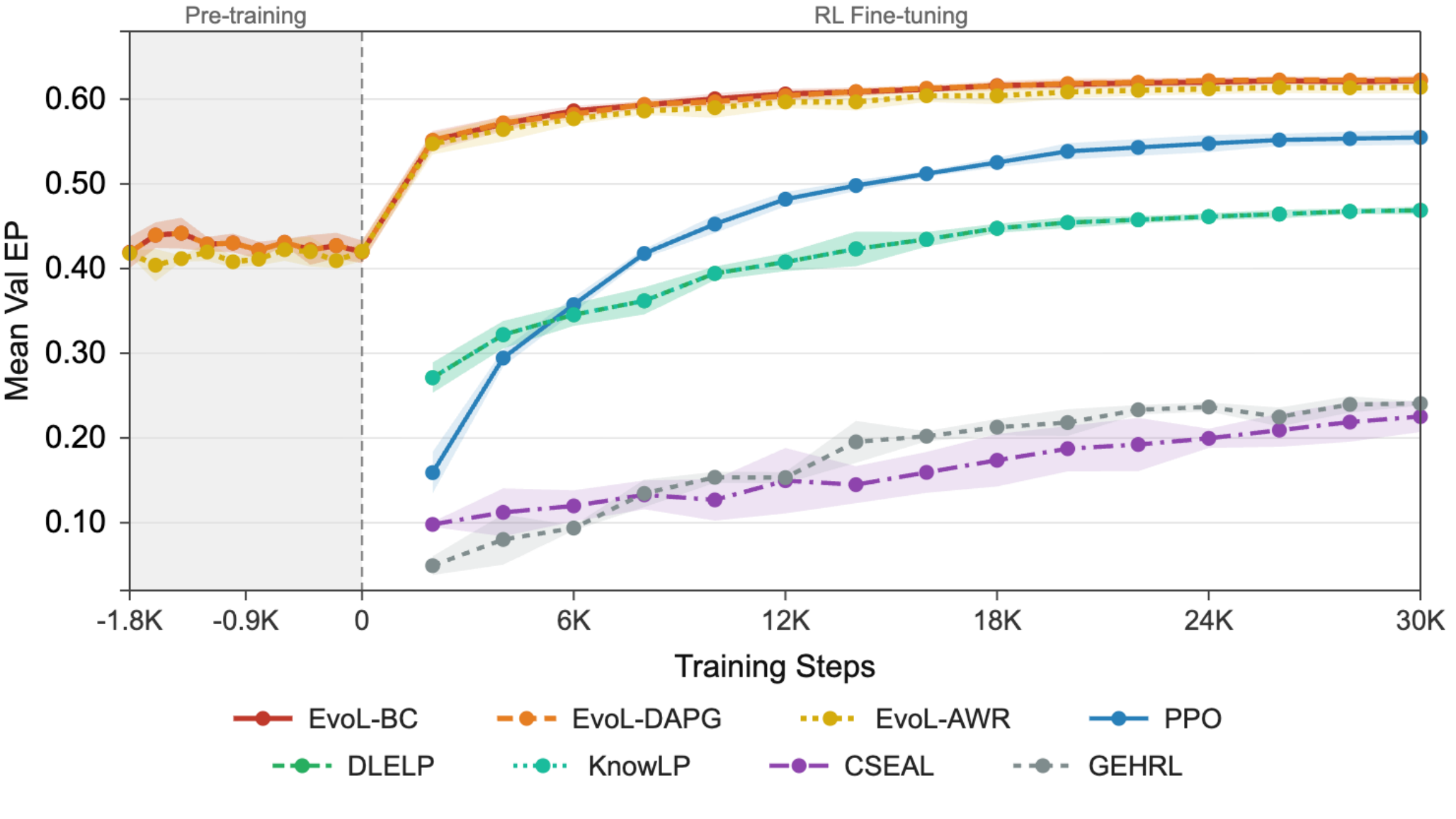}
    \caption{Validation EP during training on ASSIST15 ($L{=}20$, 3-seed mean). The \ours{} variants emerge from Stage~2 already at Val EP $\approx +0.42$; PPO-vanilla starts at $0$ and crosses the same level only around episode 6{,}000 before plateauing near $+0.51$. \ours{} continues to climb to $+0.62$ over the remaining 24{,}000 episodes.}
    \label{fig:convergence}
\end{figure}

\subsection{Component Ablations}
\label{sec:ablation}

To pinpoint which components of \ours{} contribute to the gain, we run three controlled ablations on ASSIST15 $L\!=\!10$ (3 seeds, mean$\pm$std), the same setting used by our mechanism (Section~\ref{sec:pedagogical}) and robustness (Sections~\ref{sec:cross_sim} and \ref{sec:stochastic}) analyses.

\begin{table}[t]
\centering
\caption{Component ablations on ASSIST15 $L{=}10$. All variants share the same actor-critic architecture and hyperparameters; only the indicated component is removed or replaced. Mean$\pm$std over 3 seeds (729 test learners each, except Greedy which is eval-only on 200 learners).}
\label{tab:ablation_components}
\begin{tabular}{@{}lc@{}}
\toprule
Variant & Test EP \\
\midrule
\emph{Search strategy at Stage~1} \\
\quad Greedy 1-step lookahead (no learning) & +0.155{\scriptsize$\pm$.029} \\
\quad Random-path experts $\to$ BC                  & +0.354{\scriptsize$\pm$.011} \\
\quad Evolutionary search $\to$ BC                  & \textbf{+0.614}{\scriptsize$\pm$.003} \\
\midrule
\emph{Downstream learner} \\
\quad PPO-vanilla (no experts, no pretraining)      & +0.543{\scriptsize$\pm$.009} \\
\quad \ours{}-BC w/o critic warm-start (Stage~2.5)  & +0.591{\scriptsize$\pm$.003} \\
\quad \ours{}-BC (full, Stages~1--3)                & \textbf{+0.614}{\scriptsize$\pm$.003} \\
\bottomrule
\end{tabular}
\end{table}

\textbf{(a) Why evolutionary search?}
We compare three strategies for producing Stage-1 paths under identical candidate sets and budgets. (i) Greedy 1-step lookahead, the cheapest non-random baseline, selects at each step the concept that maximizes immediate EP improvement; it ignores joint structure and reaches only $+0.155$, far below PPO-vanilla. (ii) Random-path experts discard search entirely; BC on them reaches $+0.354$, well below PPO-vanilla ($+0.543$), so any source of demonstrations is not enough---bad demonstrations actively bias the policy. (iii) Evolutionary search, which performs non-myopic global selection over the joint $L$-step space within the same candidate set, reaches $+0.614$. The contrast between greedy and evolutionary ($+0.459$ EP) is particularly informative: it isolates the value of joint sequence optimization over greedy local choices, which is precisely the combinatorial structure that motivates LPR.

\textbf{(b) Without critic warm-start.}
Removing Stage~2.5 and letting PPO learn the critic from scratch produces a small but consistent drop across all three seeds (Table~\ref{tab:ablation_components}). The warm-started critic primarily stabilizes the early phase of PPO by providing reliable advantage estimates from episode one; the asymptotic improvement is therefore expected to be modest, while the early-phase stability is the practical benefit.

\textbf{(c) Imitation algorithm is interchangeable.}
Table~\ref{tab:main} shows that BC, AWR, and DAPG yield within 0.01 EP of each other across all nine $\langle\text{dataset},L\rangle$ cells. Combined with (a), this rules out any specific imitation algorithm as the source of the improvement and confirms that evolutionary expert quality is the operative variable.

Together, the three ablations form a clean causal story: BC alone is not enough (b), random experts are not enough (a), but evolutionary experts plus a warm-started asymmetric actor-critic together produce the full effect. No single component can be dropped without measurable loss.

%% ============================================================
\section{Analysis and Discussion}
%% ============================================================

\subsection{Where the Improvement Comes From}

Table~\ref{tab:main} establishes that \ours{} surpasses every baseline in every condition, but the more interesting question is why. We answer it with three complementary observations.

\textbf{(i) Search space reduction.}
The number of distinct $L$-step paths is $\binom{M}{L}\cdot L!$, which grows rapidly with both the number of concepts and the path length, as shown in Table~\ref{tab:search_space}.

\begin{table}[t]
\centering
\caption{Combinatorial search space size across datasets and path lengths.}
\label{tab:search_space}
\normalsize
\begin{tabular}{lrrr}
\toprule
$L$ & Junyi ($M{=}39$) & ASSIST15 ($M{=}100$) & EdNet ($M{=}189$) \\
\midrule
5  & $6.9\!\times\!10^{7}$  & $9.0\!\times\!10^{9}$  & $2.3\!\times\!10^{11}$ \\
10 & $2.3\!\times\!10^{15}$ & $6.3\!\times\!10^{19}$ & $4.6\!\times\!10^{22}$ \\
20 & $1.7\!\times\!10^{29}$ & $1.3\!\times\!10^{39}$ & $1.2\!\times\!10^{45}$ \\
\bottomrule
\end{tabular}
\end{table}

RL from scratch must explore this space by sampling from a uniform initial policy, receiving non-zero reward only when the entire $L$-step path happens to improve the targets. Evolutionary search, by contrast, evaluates complete paths and selects on fitness, so it locates high-EP regions inside the simulator using $N_{\text{pop}}\!\times\!N_{\text{gen}}\!=\!3{,}000$ evaluations per learner, irrespective of $M$ or $L$. Behavioral cloning then transfers the located regions to a feed-forward policy.

\textbf{(ii) Higher starting point, shorter trajectory.}
Figure~\ref{fig:convergence} makes the optimization effect concrete. The BC-initialized \ours{} variants enter PPO from a much stronger policy than PPO-vanilla, which must spend a substantial portion of its early training budget rediscovering useful path structure from sparse terminal rewards. Even after extended fine-tuning, PPO-vanilla plateaus below \ours{}, suggesting that evolutionary pre-exploration improves not only optimization speed but also the final region reached by policy search.

\textbf{(iii) Mitigated distribution shift.}
Pure BC suffers from compounding error~\cite{ross2011dagger}: small action mistakes drive the policy into states absent from the demonstrations. In \ours{}, PPO fine-tuning explicitly trains on the policy's own state distribution, closing the gap that BC leaves open. The asymmetric critic accelerates this correction because it observes the privileged $\mathbf{h}_t$ and produces lower-variance advantage estimates, despite the actor being constrained to the deployment-realistic $\mathbf{h}_0$.

\subsection{Expert Quality Matters More Than the Imitation Objective}

A natural concern is whether the gain comes from the downstream imitation objective or from the quality of the synthesized experts. Table~\ref{tab:main} answers this directly: \ours{}-BC, \ours{}-AWR, and \ours{}-DAPG produce nearly indistinguishable final EP (within 0.01 across all nine $\langle\text{dataset},L\rangle$ conditions). AWR's advantage weighting and DAPG's decaying BC loss add no measurable benefit over plain BC.

To verify that the gains come specifically from evolutionary experts rather than from any source of demonstrations, we run an additional ablation in which the evolutionary path is replaced by a random permutation of $L$ distinct concepts (Section~\ref{sec:ablation}). All other components are kept identical. If imitation alone were enough, this random-expert variant would still benefit from BC initialization; the empirical result is the opposite.

\subsection{Pedagogical Coherence: Coverage and Sequencing}
\label{sec:pedagogical}

The improvement of \ours{} over PPO-vanilla can be decomposed along two pedagogically meaningful axes: what concepts are recommended and in what order they are arranged. To quantify these axes, we measure target coverage, defined as the fraction of target concepts appearing in the recommended path, and sequencing impact, defined as the drop in EP after uniformly shuffling the recommended path. The latter directly probes whether a path's effectiveness depends on ordering, which is a hallmark of pedagogically meaningful sequencing. Table~\ref{tab:pedagogical} reports the resulting decomposition.

\begin{table}[t]
\centering
\caption{Pedagogical decomposition on ASSIST15 $L\!=\!10$ (3 seeds, 200 test learners). ``EP$_\text{shuf}$'' is the average EP after uniformly shuffling each recommended path (5 shuffles per path); ``SeqDrop'' = EP$-$EP$_\text{shuf}$ measures the portion of EP that depends on \emph{ordering}; ``StepLift'' is the average per-step EP gain (EP$/L$).}
\label{tab:pedagogical}
\small
\begin{tabular}{@{}lccccc@{}}
\toprule
Method & EP & Coverage $\uparrow$ & EP$_\text{shuf}$ & SeqDrop $\uparrow$ & StepLift $\uparrow$ \\
\midrule
CSEAL          & +0.220 & 0.625 & +0.063 & +0.157 & +0.010 \\
GEHRL          & +0.252 & 0.784 & +0.070 & +0.182 & +0.011 \\
DLELP          & +0.479 & 0.435 & +0.213 & +0.266 & +0.017 \\
PPO-vanilla    & +0.561 & 0.713 & +0.333 & +0.227 & +0.020 \\
\midrule
\ours{}-BC     & \textbf{+0.627} & \textbf{0.876} & +0.330 & +0.297 & \textbf{+0.022} \\
\ours{}-AWR    & +0.616          & 0.864          & +0.327 & +0.289 & +0.022 \\
\ours{}-DAPG   & +0.625          & 0.861          & +0.327 & \textbf{+0.298} & +0.022 \\
\bottomrule
\end{tabular}
\end{table}

\textbf{Coverage.}
\ours{} variants cover target concepts more reliably than PPO-vanilla and the graph-enhanced methods. Coverage alone is not sufficient for high EP, since a path can include the right concepts in a poor order, but it is a basic pedagogical requirement: the recommended sequence should expose the learner to the intended goals.

\textbf{Sequencing explains most of the \ours{}--PPO gap.}
After uniformly shuffling each recommended path, \ours{}-BC and PPO-vanilla obtain nearly indistinguishable EP. This suggests that the two methods select concept sets of comparable quality, but differ in how those concepts are ordered. The original ordered paths retain a clear advantage for \ours{}, indicating that evolutionary expert distillation primarily transfers an order-sensitive policy rather than merely selecting easier concepts. Graph-enhanced baselines show a different failure mode: their shuffled paths already start from much lower EP, suggesting weaker concept selection in addition to weaker ordering.

This pattern aligns with the credit-assignment difficulty of sparse-reward RL. With only a terminal reward at step $L$, gradient updates provide little direct signal about which intermediate ordering decisions caused the final mastery improvement. Evolutionary search bypasses this difficulty by evaluating complete trajectories end-to-end, so the synthesized experts encode order-sensitive structure that the policy can absorb through imitation.

\subsection{Robustness to Simulator-Instance Shift}
\label{sec:cross_sim}

A natural concern with simulator-based training is that the learned policy might memorize idiosyncrasies of the specific KT instance used during training. To probe this, we keep each policy (and the state estimator, the original DKT trained with seed 42) fixed, and execute the recommended path against four alternative DKTs trained from scratch with seeds $\{7, 100, 200, 333 \}$ on different training subsets of $\mathcal{D}_{\text{Sim}}$. All alternative DKTs share the same architecture and hyperparameters (emb 200, dropout 0.1) as the original and reach comparable validation AUC ($0.700$--$0.708$ vs $0.704$), but encode different per-concept mastery dynamics (per-student Spearman rank correlation $0.62$). We restrict this stress test to within-DKT-family shift, i.e., alternative DKT instances trained with different random seeds and training subsets. This provides a controlled way to measure robustness to simulator variation caused by training data and initialization, which is the most direct source of simulator sensitivity in the standard KES-based LPR protocol.
\begin{table*}[t]
\centering
\caption{Robustness to DKT-instance shift on ASSIST15 $L\!=\!10$. Policies (3 seeds) trained against the original DKT (seed 42) are evaluated by executing their recommended paths on four alternative DKT instances with different training seeds (same architecture and hyperparameters). The state estimator is the original DKT throughout; only the execution simulator changes. All entries are means across 3 policy seeds $\times$ 200 test learners (same subset and evaluator as Table~\ref{tab:stochastic}, so the Original DKT column numbers differ from Table~\ref{tab:main} (729 learners) by at most $\pm 0.02$ EP per method). Methods are grouped by paradigm and, within each group, sorted by the cross-instance mean (descending). \textbf{Bold} marks the overall best per column.}
\label{tab:cross_sim}
\normalsize
\setlength{\tabcolsep}{5pt}
\begin{tabular}{@{}lccccccc@{}}
\toprule
& & \multicolumn{4}{c}{\emph{Alternative DKT (different training seed)}} & & \\
\cmidrule(lr){3-6}
Method & Original DKT & seed 7 & seed 100 & seed 200 & seed 333 & \textbf{Mean (4 alts)} & $\Delta$ vs PPO \\
\midrule
\multicolumn{8}{@{}l}{\textit{\textbf{Simulator-guided demonstration learning (Ours)}}} \\
\quad \ours{}-BC   & \textbf{+0.630} & \textbf{+0.314} & \textbf{+0.337} & \textbf{+0.350} & +0.364          & \textbf{+0.341} & \textbf{+0.091} \\
\quad \ours{}-DAPG & +0.627          & +0.306          & +0.333          & +0.346          & +0.362          & +0.337          & +0.087 \\
\quad \ours{}-AWR  & +0.619          & +0.309          & +0.330          & +0.339          & \textbf{+0.367} & +0.336          & +0.086 \\
\midrule
\multicolumn{8}{@{}l}{\textit{Reinforcement learning baseline}} \\
\quad PPO-vanilla  & +0.558          & +0.219          & +0.270          & +0.240          & +0.269          & +0.250          & ---     \\
\midrule
\multicolumn{8}{@{}l}{\textit{Sequence recommenders}} \\
\quad SASRec       & +0.483          & +0.244          & +0.180          & +0.246          & +0.230          & +0.225          & $-$0.025 \\
\quad GRU4Rec      & +0.330          & +0.106          & +0.034          & +0.117          & +0.095          & +0.088          & $-$0.162 \\
\midrule
\multicolumn{8}{@{}l}{\textit{Knowledge-graph-based RL}} \\
\quad DLELP        & +0.471          & +0.258          & +0.184          & +0.212          & +0.206          & +0.215          & $-$0.035 \\
\quad GEHRL        & +0.246          & +0.045          & +0.001          & +0.054          & +0.035          & +0.034          & $-$0.216 \\
\quad CSEAL        & +0.211          & +0.057          & $-$0.020        & +0.053          & +0.017          & +0.027          & $-$0.223 \\
\midrule
\multicolumn{8}{@{}l}{\textit{LLM-enhanced graph-based RL}} \\
\quad KnowLP       & +0.299          & +0.134          & +0.141          & +0.191          & +0.141          & +0.152          & $-$0.098 \\
\bottomrule
\end{tabular}
\\[2pt]
{\footnotesize $\Delta$ vs PPO: difference in cross-instance mean against PPO-vanilla.}
\end{table*}

Several patterns emerge from Table~\ref{tab:cross_sim}. First, all three \ours{} variants remain the strongest methods across the alternative DKT instances, and their differences are much smaller than the gap to the strongest baseline. This supports the view that the robust component is evolutionary expert synthesis rather than the choice of BC, AWR, or DAPG. Second, PPO-vanilla and sequence models retain part of their performance, whereas graph-conditioned baselines degrade more sharply, suggesting that fixed graph-derived candidate structures are more sensitive to simulator-instance changes. Finally, the absolute drop observed across methods confirms that DKT-instance shift is a meaningful stress test rather than a benign re-evaluation setting.

\subsection{Robustness to Stochastic Response Models}
\label{sec:stochastic}

A second concern specific to \ours{} is that Stage~1 optimizes EP directly inside the simulator, so the evolutionary search might exploit discontinuities introduced by the deterministic threshold $r\!=\!\mathbf{1}[h_c\!>\!0.5]$. We re-evaluate every trained policy (deterministic-trained checkpoints; no re-training) under three alternative response models that smooth out the threshold:
(i) \emph{Bernoulli}, $r\!\sim\!\mathrm{Bern}(h_c)$;
(ii) \emph{Slip/Guess (mild)}, $P(r{=}1)=(1{-}s)h_c + g(1{-}h_c)$ with $s\!=\!g\!=\!0.10$;
(iii) \emph{Slip/Guess (strong)}, the same with $s\!=\!g\!=\!0.20$.
Paths are deterministic ($\mathbf{h}_0$-only actors); only the response is randomized, with 20 stochastic rollouts per learner to reduce variance. The same perturbation is applied uniformly to all seven non-ours baselines (PPO-vanilla, SASRec, GRU4Rec, DLELP, KnowLP, GEHRL, CSEAL); the deterministic reference column uses the same 200-learner subset and the same evaluator code-path, so numbers differ from Table~\ref{tab:main} (729 learners) by at most $\pm 0.02$ EP.

\begin{table*}[t]
\centering
\caption{Robustness to stochastic response models on ASSIST15 $L\!=\!10$. The deterministic-trained checkpoint of each method is evaluated under Bernoulli and slip/guess response models without re-training (3 policy seeds $\times$ 200 test learners $\times$ 20 stochastic rollouts; $\mathbf{h}_0$-only paths). Methods grouped by paradigm and sorted within group by the mean across the three stochastic modes. \textbf{Bold} marks the overall best per column.}
\label{tab:stochastic}
\normalsize
\setlength{\tabcolsep}{5pt}
\begin{tabular}{@{}lcccccc@{}}
\toprule
& & \multicolumn{3}{c}{\emph{Stochastic response model}} & & \\
\cmidrule(lr){3-5}
Method & Determ.\ (ref.) & Bernoulli & Slip/Guess (0.10) & Slip/Guess (0.20) & \textbf{Mean (3 stoch.)} & $\Delta$ vs PPO \\
\midrule
\multicolumn{7}{@{}l}{\textit{\textbf{Simulator-guided demonstration learning (Ours)}}} \\
\quad \ours{}-BC   & \textbf{+0.630} & \textbf{+0.305} & \textbf{+0.272} & +0.223          & \textbf{+0.267} & \textbf{+0.043} \\
\quad \ours{}-DAPG & +0.627          & +0.303          & +0.270          & \textbf{+0.224} & +0.266          & +0.042 \\
\quad \ours{}-AWR  & +0.619          & +0.302          & \textbf{+0.272} & +0.221          & +0.265          & +0.041 \\
\midrule
\multicolumn{7}{@{}l}{\textit{Reinforcement learning baseline}} \\
\quad PPO-vanilla  & +0.558          & +0.267          & +0.227          & +0.179          & +0.224          & ---     \\
\midrule
\multicolumn{7}{@{}l}{\textit{Sequence recommenders}} \\
\quad SASRec       & +0.483          & +0.146          & +0.125          & +0.094          & +0.122          & $-$0.102 \\
\quad GRU4Rec      & +0.330          & +0.009          & $-$0.014        & $-$0.039        & $-$0.015        & $-$0.239 \\
\midrule
\multicolumn{7}{@{}l}{\textit{Knowledge-graph-based RL}} \\
\quad DLELP        & +0.471          & +0.153          & +0.118          & +0.074          & +0.115          & $-$0.109 \\
\quad GEHRL        & +0.246          & $-$0.093        & $-$0.106        & $-$0.135        & $-$0.111        & $-$0.335 \\
\quad CSEAL        & +0.211          & $-$0.088        & $-$0.107        & $-$0.133        & $-$0.109        & $-$0.333 \\
\midrule
\multicolumn{7}{@{}l}{\textit{LLM-enhanced graph-based RL}} \\
\quad KnowLP       & +0.299          & $-$0.013        & $-$0.042        & $-$0.082        & $-$0.046        & $-$0.270 \\
\bottomrule
\end{tabular}
\\[2pt]
{\footnotesize $\Delta$ vs PPO: difference between the method's mean across the three stochastic modes and PPO-vanilla's mean. Bold entries identify the column-wise best across all ten methods.}
\end{table*}

Three observations are evident from Table~\ref{tab:stochastic}. First, all three \ours{} variants remain the strongest methods under every stochastic response model, which weakens the hypothesis that evolutionary search merely exploits deterministic threshold discontinuities. Second, all methods become harder to optimize under stochastic responses, but the relative advantage of \ours{} is preserved. Third, graph-conditioned baselines are more fragile under response perturbations, suggesting that policies whose action spaces are strongly shaped by fixed concept graphs may be more sensitive to changes in simulator dynamics.

Together, the simulator-instance and response-model stress tests show that \ours{}'s advantage is not tied to a single deterministic DKT instance.

\subsection{Generalizability of the Recipe}

The proposed recipe is not tied to a particular dataset or imitation objective.
It requires three ingredients: (i) a simulator that can be queried during
training, (ii) a combinatorial sequence decision problem with sparse terminal
reward, and (iii) instance-specific near-optimal sequences that human
demonstrations cannot supply. Learning path recommendation satisfies all three:
the KES evaluates candidate paths offline, the path count grows combinatorially
with $M$ and $L$, and student logs record historical behavior rather than expert
prescriptions.

Within this setting, \ours{} separates simulator use from deployment: the
simulator synthesizes high-quality per-learner expert paths offline, and the
resulting behavior is distilled into an actor that consumes only deployable
inputs. This separation is the central recipe---expensive or privileged search
during training, but a deployed policy that acts without simulator rollouts or
counterfactual intermediate feedback. The same pattern applies wherever
training-time simulators exist but deployment-time interaction must be planned
in advance, such as personalized tutoring across curricula, treatment-sequence
recommendation with patient simulators, or content-layout planning with
user-response simulators. Security-oriented settings exhibit a closely related
structure. Attack-path or defense-strategy planning can be rehearsed inside a
cyber-twin replica of a target network before any action touches the live
system, and multi-step tracing over heterogeneous evidence graphs in digital
investigations, which link anonymized network nodes, asset flows, and online
advertisements, yields a useful outcome only once the full chain of probes is
resolved. Both face the sparse and order-sensitive terminal signal that
motivates \ours{}.

These claims are scoped to simulator-based policy learning rather than to
measured real-world learning gains, and two limitations follow. First, expert
synthesis optimizes EP inside a DKT-based simulator, so the demonstrations
inherit whatever the DKT fails to represent---most notably, two learners with
identical histories are assigned identical dynamics regardless of differing
aptitude; the robustness analyses above reduce the risk of exploiting a single
DKT instance, but prospective validation with live learners and human-designed
curriculum constraints remains an important next step. Second, baselines
originally designed for inference-time simulator querying are evaluated under
the stricter $\mathbf{h}_0$-only constraint, so the reported gaps characterize
deployment-free performance rather than the original methods under their own
assumptions. Neither affects the main comparison, where all methods face the
same deployment-free information constraint.

%% ============================================================
\section{Conclusion}
%% ============================================================

We presented \ours{}, a simulator-guided demonstration-learning framework for deployment-free learning path recommendation. \ours{} uses a DKT-based knowledge evolution simulator not only as a training environment but also as a source of per-learner expert paths synthesized through evolutionary search. These demonstrations are distilled into an asymmetric actor-critic whose actor plans from the initial mastery state alone, enabling KES-free recommendation at deployment. Across datasets, path lengths, and imitation variants, the results indicate that the central bottleneck in deployment-realistic LPR is not the imitation objective itself, but the synthesis of high-quality expert trajectories that can guide sparse-reward policy learning.

\section*{Acknowledgement}
This research was supported by Development of Field Technology for Responding to Illegal Drugs Program through the Korea Institutes of Police Technology (KIPoT) funded by the Ministry of Science and ICT \& Korean National Police Agency (RS-2026-25535833), and this research (Development of a zero-hacking system based on AI white-hat hackers) was supported by Korea Institute of Science and Technology Information (KISTI)(P26035, K26L8M1C1). This research was supported by the MSIT(Ministry of Science and ICT), Korea, under the Graduate School of Virtual Convergence support program(IITP-2026-RS-2023-00254129) supervised by the IITP(Institute for Information \& Communications Technology Planning \& Evaluation).

\section*{GenAI Usage Disclosure}

The authors used LLM-based tools only for writing assistance, including English drafting, grammar correction, and wording refinement. All AI-assisted content was reviewed and edited by the authors, who take full responsibility for the final manuscript.

\bibliographystyle{ACM-Reference-Format}
\balance
\bibliography{reference}

@inproceedings{piech2015dkt,
 author = {Piech, Chris and Bassen, Jonathan and Huang, Jonathan and Ganguli, Surya and Sahami, Mehran and Guibas, Leonidas and Sohl-Dickstein, Jascha},
 booktitle = {Advances in Neural Information Processing Systems},
 pages = {},
 publisher = {Curran Associates, Inc.},
 title = {Deep Knowledge Tracing},
 url = {https://proceedings.neurips.cc/paper_files/paper/2015/file/bac9162b47c56fc8a4d2a519803d51b3-Paper.pdf},
 volume = {28},
 year = {2015}
}

@inproceedings{liu2019cseal,
author = {Liu, Qi and Tong, Shiwei and Liu, Chuanren and Zhao, Hongke and Chen, Enhong and Ma, Haiping and Wang, Shijin},
title = {Exploiting Cognitive Structure for Adaptive Learning},
year = {2019},
isbn = {9781450362016},
publisher = {Association for Computing Machinery},
address = {New York, NY, USA},
url = {https://doi.org/10.1145/3292500.3330922},
doi = {10.1145/3292500.3330922},
booktitle = {Proceedings of the 25th ACM SIGKDD International Conference on Knowledge Discovery \& Data Mining},
pages = {627–635},
numpages = {9},
location = {Anchorage, AK, USA},
series = {KDD '19}
}

@inproceedings{li2023gehrl,
author = {Li, Qingyao and Xia, Wei and Yin, Li'ang and Shen, Jian and Rui, Renting and Zhang, Weinan and Chen, Xianyu and Tang, Ruiming and Yu, Yong},
title = {Graph Enhanced Hierarchical Reinforcement Learning for Goal-oriented Learning Path Recommendation},
year = {2023},
isbn = {9798400701245},
publisher = {Association for Computing Machinery},
address = {New York, NY, USA},
url = {https://doi.org/10.1145/3583780.3614897},
doi = {10.1145/3583780.3614897},
booktitle = {Proceedings of the 32nd ACM International Conference on Information and Knowledge Management},
pages = {1318–1327},
numpages = {10},
location = {Birmingham, United Kingdom},
series = {CIKM '23}
}

@inproceedings{cheng2025knowlp,
  title={GraphRAG-Induced Dual Knowledge Structure Graphs for Personalized Learning Path Recommendation},
  author={Cheng, Xinghe and Zhang, Zihan and Wang, Jiapu and Fang, Liangda and He, Chaobo and Guan, Quanlong and Pan, Shirui and Luo, Weiqi},
  booktitle={Proceedings of the AAAI Conference on Artificial Intelligence},
  volume={40},
  number={17},
  pages={14610--14620},
  year={2026}
}

@inproceedings{chen2023src,
  title={Set-to-Sequence Ranking-Based Concept-Aware Learning Path Recommendation},
  author={Chen, Xianyu and Shen, Jian and Xia, Wei and Jin, Jiarui and Song, Yakun and Zhang, Weinan and Liu, Weiwen and Zhu, Menghui and Tang, Ruiming and Dong, Kai and others},
  booktitle={Proceedings of the AAAI Conference on Artificial Intelligence},
  volume={37},
  number={4},
  pages={5027--5035},
  year={2023}
}

@article{sweller1988cognitive,
title = {Cognitive load during problem solving: Effects on learning},
journal = {Cognitive Science},
volume = {12},
number = {2},
pages = {257-285},
year = {1988},
issn = {0364-0213},
doi = {https://doi.org/10.1016/0364-0213(88)90023-7},
url = {https://www.sciencedirect.com/science/article/pii/0364021388900237},
author = {John Sweller},
}

@misc{schulman2017ppo,
      title={Proximal Policy Optimization Algorithms}, 
      author={John Schulman and Filip Wolski and Prafulla Dhariwal and Alec Radford and Oleg Klimov},
      year={2017},
      eprint={1707.06347},
      archivePrefix={arXiv},
      primaryClass={cs.LG},
      url={https://arxiv.org/abs/1707.06347}, 
}

@article{sweller1998cognitive,
  title={Cognitive Architecture and Instructional Design
},
  author={Sweller, John and Van Merrienboer, Jeroen JG and Paas, Fred GWC},
  journal={Educational Psychology Review},
  volume={10},
  number={3},
  pages={251--296},
  year={1998},
  publisher={Springer},
  doi={https://doi.org/10.1023/A:1022193728205}
}

@misc{hidasi2016gru4rec,
      title={Session-based Recommendations with Recurrent Neural Networks}, 
      author={Balázs Hidasi and Alexandros Karatzoglou and Linas Baltrunas and Domonkos Tikk},
      year={2016},
      eprint={1511.06939},
      archivePrefix={arXiv},
      primaryClass={cs.LG},
      url={https://arxiv.org/abs/1511.06939}, 
}

@inproceedings{zhang2017dkvmn,
author = {Zhang, Jiani and Shi, Xingjian and King, Irwin and Yeung, Dit-Yan},
title = {Dynamic Key-Value Memory Networks for Knowledge Tracing},
year = {2017},
isbn = {9781450349130},
publisher = {International World Wide Web Conferences Steering Committee},
address = {Republic and Canton of Geneva, CHE},
url = {https://doi.org/10.1145/3038912.3052580},
doi = {10.1145/3038912.3052580},
booktitle = {Proceedings of the 26th International Conference on World Wide Web},
pages = {765–774},
numpages = {10},
location = {Perth, Australia},
series = {WWW '17}
}

@inproceedings{pandey2019sakt,
  author    = {Shalini Pandey and George Karypis},
  title     = {A Self-Attentive Model for Knowledge Tracing},
  booktitle = {Proceedings of the 12th International Conference on Educational Data Mining (EDM 2019)},
  year      = {2019}
}

@inproceedings{ghosh2020akt,
author = {Ghosh, Aritra and Heffernan, Neil and Lan, Andrew S.},
title = {Context-Aware Attentive Knowledge Tracing},
year = {2020},
isbn = {9781450379984},
publisher = {Association for Computing Machinery},
address = {New York, NY, USA},
url = {https://doi.org/10.1145/3394486.3403282},
doi = {10.1145/3394486.3403282},
booktitle = {Proceedings of the 26th ACM SIGKDD International Conference on Knowledge Discovery \& Data Mining},
pages = {2330–2339},
numpages = {10},
location = {Virtual Event, CA, USA},
series = {KDD '20}
}

@inproceedings{choi2020saint,
author = {Choi, Youngduck and Lee, Youngnam and Cho, Junghyun and Baek, Jineon and Kim, Byungsoo and Cha, Yeongmin and Shin, Dongmin and Bae, Chan and Heo, Jaewe},
title = {Towards an Appropriate Query, Key, and Value Computation for Knowledge Tracing},
year = {2020},
isbn = {9781450379519},
publisher = {Association for Computing Machinery},
address = {New York, NY, USA},
url = {https://doi.org/10.1145/3386527.3405945},
doi = {10.1145/3386527.3405945},
booktitle = {Proceedings of the Seventh ACM Conference on Learning @ Scale},
pages = {341–344},
numpages = {4},
location = {Virtual Event, USA},
series = {L@S '20}
}

@InProceedings{choi2020ednet,
author="Choi, Youngduck and Lee, Youngnam and Shin, Dongmin and Cho, Junghyun and Park, Seoyon and Lee, Seewoo and Baek, Jineon and Bae, Chan and Kim, Byungsoo and Heo, Jaewe",
title="EdNet: A Large-Scale Hierarchical Dataset in Education",
booktitle="Artificial Intelligence in Education",
year="2020",
publisher="Springer International Publishing",
address="Cham",
pages="69--73",
isbn="978-3-030-52240-7"
}

@article{feng2009assistments,
  title={Addressing the assessment challenge with an online system that tutors as it assesses},
  author={Feng, Mingyu and Heffernan, Neil and Koedinger, Kenneth},
  journal={User Modeling and User-Adapted Interaction},
  volume={19},
  number={3},
  pages={243--266},
  year={2009},
  publisher={Springer},
  doi={https://doi.org/10.1007/s11257-009-9063-7}
}

@inproceedings{chang2015junyi,
  author = {Chang, Haw{-}Shiuan and Hsu, Hwai{-}Jung and Chen, Kuan{-}Ta},
  title = {Modeling Exercise Relationships in E-Learning: A Unified Approach},
  booktitle = {International Conference on Educational Data Mining (EDM)},
  year = {2015}
}

@INPROCEEDINGS{chen2018prerequisite,
  author={Chen, Penghe and Lu, Yu and Zheng, Vincent W. and Pian, Yang},
  booktitle={2018 IEEE International Conference on Data Mining (ICDM)}, 
  title={Prerequisite-Driven Deep Knowledge Tracing}, 
  year={2018},
  volume={},
  number={},
  pages={39-48},
  doi={10.1109/ICDM.2018.00019}
}

@InProceedings{ross2011dagger,
  title = 	 {A Reduction of Imitation Learning and Structured Prediction to No-Regret Online Learning},
  author = 	 {Ross, Stephane and Gordon, Geoffrey and Bagnell, Drew},
  booktitle = 	 {Proceedings of the Fourteenth International Conference on Artificial Intelligence and Statistics},
  pages = 	 {627--635},
  year = 	 {2011},
  volume = 	 {15},
  series = 	 {Proceedings of Machine Learning Research},
  address = 	 {Fort Lauderdale, FL, USA},
  month = 	 {11--13 Apr},
  publisher =    {PMLR},
  url = 	 {https://proceedings.mlr.press/v15/ross11a.html},
}

@inproceedings{torabi2018bc,
  title     = {Behavioral Cloning from Observation},
  author    = {Faraz Torabi and Garrett Warnell and Peter Stone},
  booktitle = {Proceedings of the Twenty-Seventh International Joint Conference on Artificial Intelligence, {IJCAI-18}},
  publisher = {International Joint Conferences on Artificial Intelligence Organization},
  pages     = {4950--4957},
  year      = {2018},
  month     = {7},
  doi       = {10.24963/ijcai.2018/687},
  url       = {https://doi.org/10.24963/ijcai.2018/687},
}

@misc{jaderberg2017pbt,
      title={Population Based Training of Neural Networks}, 
      author={Max Jaderberg and Valentin Dalibard and Simon Osindero and Wojciech M. Czarnecki and Jeff Donahue and Ali Razavi and Oriol Vinyals and Tim Green and Iain Dunning and Karen Simonyan and Chrisantha Fernando and Koray Kavukcuoglu},
      year={2017},
      eprint={1711.09846},
      archivePrefix={arXiv},
      primaryClass={cs.LG},
      url={https://arxiv.org/abs/1711.09846}, 
}

@misc{salimans2017es,
      title={Evolution Strategies as a Scalable Alternative to Reinforcement Learning}, 
      author={Tim Salimans and Jonathan Ho and Xi Chen and Szymon Sidor and Ilya Sutskever},
      year={2017},
      eprint={1703.03864},
      archivePrefix={arXiv},
      primaryClass={stat.ML},
      url={https://arxiv.org/abs/1703.03864}, 
}

@inproceedings{khadka2018erl,
 author = {Khadka, Shauharda and Tumer, Kagan},
 booktitle = {Advances in Neural Information Processing Systems},
 pages = {},
 publisher = {Curran Associates, Inc.},
 title = {Evolution-Guided Policy Gradient in Reinforcement Learning},
 url = {https://proceedings.neurips.cc/paper_files/paper/2018/file/85fc37b18c57097425b52fc7afbb6969-Paper.pdf},
 volume = {31},
 year = {2018}
}

@article{corbett1994knowledge,
  title={Knowledge tracing: Modeling the acquisition of procedural knowledge},
  author={Corbett, Albert T and Anderson, John R},
  journal={User modeling and user-adapted interaction},
  volume={4},
  number={4},
  pages={253--278},
  year={1994},
  publisher={Springer},
  doi={https://doi.org/10.1007/BF01099821}
}

@ARTICLE{shen2024survey,
  author={Shen, Shuanghong and Liu, Qi and Huang, Zhenya and Zheng, Yonghe and Yin, Minghao and Wang, Minjuan and Chen, Enhong},
  journal={IEEE Transactions on Learning Technologies}, 
  title={A Survey of Knowledge Tracing: Models, Variants, and Applications}, 
  year={2024},
  volume={17},
  number={},
  pages={1858-1879},
  doi={10.1109/TLT.2024.3383325}}

@INPROCEEDINGS{yamada2024twist,
  author={Yamada, Jun and Rigter, Marc and Collins, Jack and Posner, Ingmar},
  booktitle={2024 IEEE International Conference on Robotics and Automation (ICRA)}, 
  title={TWIST: Teacher-Student World Model Distillation for Efficient Sim-to-Real Transfer}, 
  year={2024},
  volume={},
  number={},
  pages={9190-9196},
  doi={10.1109/ICRA57147.2024.10610450}}

@INPROCEEDINGS{rajeswaran2018dexterous,
    AUTHOR = {Aravind Rajeswaran AND Vikash Kumar AND Abhishek Gupta AND Giulia Vezzani AND John Schulman AND Emanuel Todorov AND Sergey Levine},
    TITLE = {Learning Complex Dexterous Manipulation with Deep Reinforcement Learning and Demonstrations},
    BOOKTITLE = {Proceedings of Robotics: Science and Systems (RSS)},
    YEAR = {2018},
}

@INPROCEEDINGS{nair2018demonstrations,
  author={Nair, Ashvin and McGrew, Bob and Andrychowicz, Marcin and Zaremba, Wojciech and Abbeel, Pieter},
  booktitle={2018 IEEE International Conference on Robotics and Automation (ICRA)}, 
  title={Overcoming Exploration in Reinforcement Learning with Demonstrations}, 
  year={2018},
  volume={},
  number={},
  pages={6292-6299},
  doi={10.1109/ICRA.2018.8463162}}

@inproceedings{lange2023evosax,
author = {Lange, Robert Tjarko},
title = {evosax: JAX-Based Evolution Strategies},
year = {2023},
isbn = {9798400701207},
publisher = {Association for Computing Machinery},
address = {New York, NY, USA},
url = {https://doi.org/10.1145/3583133.3590733},
doi = {10.1145/3583133.3590733},
booktitle = {Proceedings of the Companion Conference on Genetic and Evolutionary Computation},
pages = {659–662},
numpages = {4},
location = {Lisbon, Portugal},
series = {GECCO '23 Companion}
}

@ARTICLE{li2024evorl,
  author={Li, Pengyi and Hao, Jianye and Tang, Hongyao and Fu, Xian and Zheng, Yan and Tang, Ke},
  journal={IEEE Transactions on Evolutionary Computation}, 
  title={Bridging Evolutionary Algorithms and Reinforcement Learning: A Comprehensive Survey on Hybrid Algorithms}, 
  year={2025},
  volume={29},
  number={5},
  pages={1707-1728},
  doi={10.1109/TEVC.2024.3443913}}

@inproceedings{li2024privileged,
author = {Li, Qingyao and Xia, Wei and Yin, Li'ang and Jin, Jiarui and Yu, Yong},
title = {Privileged Knowledge State Distillation for Reinforcement Learning-based Educational Path Recommendation},
year = {2024},
isbn = {9798400704901},
publisher = {Association for Computing Machinery},
address = {New York, NY, USA},
url = {https://doi.org/10.1145/3637528.3671872},
doi = {10.1145/3637528.3671872},
booktitle = {Proceedings of the 30th ACM SIGKDD Conference on Knowledge Discovery and Data Mining},
pages = {1621–1630},
numpages = {10},
location = {Barcelona, Spain},
series = {KDD '24}
}

@misc{peng2019awr,
      title={Advantage-Weighted Regression: Simple and Scalable Off-Policy Reinforcement Learning}, 
      author={Xue Bin Peng and Aviral Kumar and Grace Zhang and Sergey Levine},
      year={2019},
      eprint={1910.00177},
      archivePrefix={arXiv},
      primaryClass={cs.LG},
      url={https://arxiv.org/abs/1910.00177}, 
}

@INPROCEEDINGS{kang2018sasrec,
  author={Kang, Wang-Cheng and McAuley, Julian},
  booktitle={2018 IEEE International Conference on Data Mining (ICDM)}, 
  title={Self-Attentive Sequential Recommendation}, 
  year={2018},
  volume={},
  number={},
  pages={197-206},
  doi={10.1109/ICDM.2018.00035}}

@article{hwangbo2019learning,
author = {Jemin Hwangbo  and Joonho Lee  and Alexey Dosovitskiy  and Dario Bellicoso  and Vassilios Tsounis  and Vladlen Koltun  and Marco Hutter },
title = {Learning agile and dynamic motor skills for legged robots},
journal = {Science Robotics},
volume = {4},
number = {26},
pages = {eaau5872},
year = {2019},
doi = {10.1126/scirobotics.aau5872}
}

@inproceedings{liu2022pykt,
 author = {Liu, Zitao and Liu, Qiongqiong and Chen, Jiahao and Huang, Shuyan and Tang, Jiliang and Luo, Weiqi},
 booktitle = {Advances in Neural Information Processing Systems},
 doi = {10.52202/068431-1347},
 pages = {18542--18555},
 publisher = {Curran Associates, Inc.},
 title = {pyKT: A Python Library to Benchmark Deep Learning based Knowledge Tracing Models},
 url = {https://proceedings.neurips.cc/paper_files/paper/2022/file/75ca2b23d9794f02a92449af65a57556-Paper-Datasets_and_Benchmarks.pdf},
 volume = {35},
 year = {2022}
}

@INPROCEEDINGS{todorov2012mujoco,
  author={Todorov, Emanuel and Erez, Tom and Tassa, Yuval},
  booktitle={2012 IEEE/RSJ International Conference on Intelligent Robots and Systems}, 
  title={MuJoCo: A physics engine for model-based control}, 
  year={2012},
  volume={},
  number={},
  pages={5026-5033},
  doi={10.1109/IROS.2012.6386109}}

@INPROCEEDINGS{peng2018sim,
  author={Peng, Xue Bin and Andrychowicz, Marcin and Zaremba, Wojciech and Abbeel, Pieter},
  booktitle={2018 IEEE International Conference on Robotics and Automation (ICRA)}, 
  title={Sim-to-Real Transfer of Robotic Control with Dynamics Randomization}, 
  year={2018},
  volume={},
  number={},
  pages={3803-3810},
  doi={10.1109/ICRA.2018.8460528}}

@misc{makoviychuk2021isaac,
      title={Isaac Gym: High Performance GPU-Based Physics Simulation For Robot Learning}, 
      author={Viktor Makoviychuk and Lukasz Wawrzyniak and Yunrong Guo and Michelle Lu and Kier Storey and Miles Macklin and David Hoeller and Nikita Rudin and Arthur Allshire and Ankur Handa and Gavriel State},
      year={2021},
      eprint={2108.10470},
      archivePrefix={arXiv},
      primaryClass={cs.RO},
      url={https://arxiv.org/abs/2108.10470}, 
}

\end{document}